\documentclass[letterpaper, 10 pt, conference]{ieeeconf}
\IEEEoverridecommandlockouts % ieeeconf: allows \thanks etc. in \author
\usepackage{cite}
\usepackage{amsmath,amssymb,amsfonts}
\usepackage{graphicx}
\let\labelindent\relax   % ieeeconf defines \labelindent; avoid the clash with enumitem
\usepackage{enumitem}
\usepackage{textcomp}
\usepackage{xcolor}
\usepackage{booktabs}
\usepackage{url}

\usepackage{caption}

\providecommand{\sd}[1]{{\scriptsize\,$\pm$#1}}

\newif\ifblind % 査読中は true，カメラレディ時に false にする
\blindfalse
\newif\ifpix % 査読中は true，カメラレディ時に false にする
\pixfalse
\begin{document}

\title{\LARGE \bf
Self-Supervised Anchoring of {Fingertip} Sensing to \\
Proprioception and Proactive Actions for Robot Imitation Learning
}

\ifblind
  \author{Anonymous Author(s)}
\else
    \IEEEoverridecommandlockouts
        \author{Tomohiro~Motoda$^{1,*}$, % 学習モデルの提案，論文の執筆
        Masaki~Murooka$^{2}$, % 議論・ハードウェア提案・設計・調達
        Keisuke~Shirai$^{1}$,
        Hanbit~Oh$^{1}$,
        Ryoichi~Nakajo$^{1}$,\\
        Shotaro~Miwa$^{1}$,
        Roman~Mykhailyshyn$^{1}$,
        Hugo~Duarte$^{1}$,
        and Yukiyasu~Domae$^{1}$
        \thanks{$^{1}$Embodied AI Research Team, AIRC and National Institute of Advanced Industrial Science and Technology (AIST)}
        \thanks{$^{2}$CNRS-AIST JRL (Joint Robotics Laboratory), IRL and National Institute of Advanced Industrial Science and Technology (AIST)}
        }
\fi

\maketitle
\thispagestyle{empty}
\pagestyle{empty}

\begin{abstract}

Robotic imitation learning often relies on external cameras, yet local interaction cues such as object proximity, contact onset, and grasp state are difficult to observe near the fingertips because of occlusion and limited temporal resolution. We study how to effectively incorporate complementary fingertip sensing into imitation learning using pressure-sensitive tactile and reflective proximity sensors, along with pretrained sensor encoders. The two modalities provide information at different manipulation phases: proximity sensing is informative before contact, whereas tactile sensing becomes informative after contact. However, naively adding these signals to a policy does not consistently improve performance and can even underperform vision-only policies, suggesting that sparse, phase-dependent sensor signals are difficult to exploit from limited demonstrations. We therefore propose a proprioception-anchored pretraining method, \underline{PRO}prioceptive-and-\underline{PR}oactive \underline{A}nchoring (PROPRA), which independently aligns each fingertip sensor history with proprioceptive and action segments. 
This provides a continuously available sensorimotor reference, allowing each sensor to be aligned independently during its informative phases. Experiments on real-world manipulation tasks show that our pretraining method improves average success rates over vision-only policies and image-anchored pretraining baselines. Representation analysis further shows that it preserves richer information about pre-contact states, enabling more effective use of complementary fingertip sensing. 
Please refer to our project page:~\url{https://tomohiromotoda.github.io/nia.propra/}

%Please refer to our project page:~\url{https://anonymous.4open.science/w/own-F865/}. 
\end{abstract}

% \begin{IEEEkeywords}
% Tactile and proximity sensing, Self-supervised representation learning, Contrastive pretraining, Proprioception, Imitation learning
% \end{IEEEkeywords}

% =============================================================================
\section{Introduction}
\label{sec:intro}

Fingertip sensing provides rich perceptual information that plays a central role in dexterous human manipulation. Humans flexibly adapt their movements to the characteristics and state of a task by integrating multiple sensory signals, including vision and touch. Even in fundamental grasping behaviors, such multimodal feedback enables us to execute precise movements and to maintain reliable grasping even when visual information is partially or temporarily unavailable. Endowing robotic systems with similar fingertip sensing capabilities is therefore a promising direction toward more dexterous and robust manipulation.

The importance of incorporating multimodal perceptual information near the fingertips into imitation learning for robotic manipulation has been increasingly recognized~\cite{gelsight, proxtac, vital}. However, different sensors provide distinct types of information, and their informativeness varies across the manipulation process. In particular, recognizing not only the post-contact state but also the \emph{pre-grasp} state, in which the robot approaches the object and establishes a graspable spatial configuration, is known to be crucial for determining the subsequent contact and grasp outcome~\cite{koyama2013preshape,looktotouch,rupavatharam2023sonicfinger}. Yet, even when complementary sensing is exploited, naively concatenating modalities can degrade the policy on some tasks, where sensor-competition effects interfere with each other~\cite{Wang2020multihard}. Importantly, the challenge extends beyond policy fusion. When sensory modalities become informative at different phases, learning representations by directly aligning modalities can also produce weak or ambiguous supervision signals. For example, a modality may provide little or no informative signal during periods when another modality is most relevant, making temporal co-occurrence an unreliable supervision cue. Because each modality becomes informative at different stages of manipulation, the learned representation should capture when each sensor contains task-relevant information and how that information relates to the robot's motion and physical state. Representation learning is therefore important for converting heterogeneous fingertip signals into features that an imitation learning policy can effectively exploit.

\begin{figure}[!t]
    \centering
    \IfFileExists{fig/top.pdf}{\includegraphics[width=0.99\columnwidth]{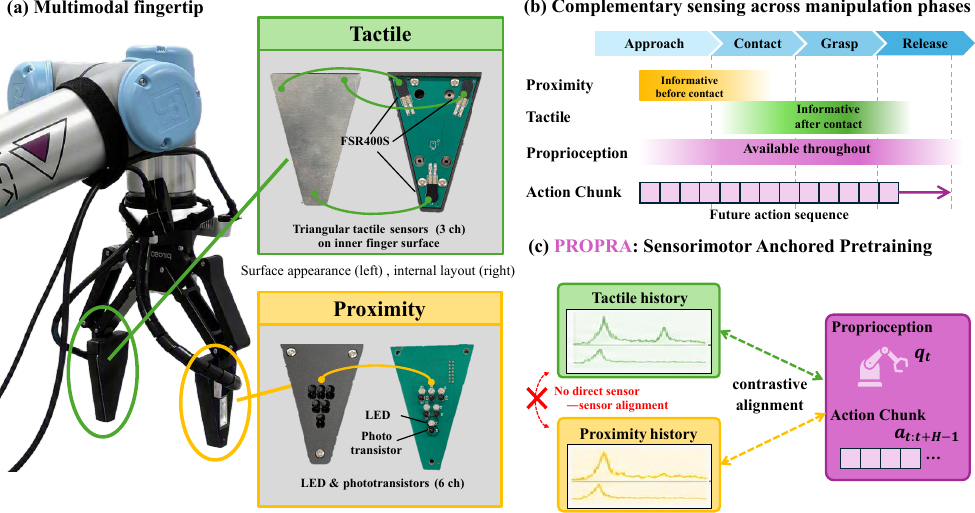}}
    {\fbox{\parbox[c][30mm][c]{0.92\columnwidth}{\centering\footnotesize
     PLACEHOLDER: \texttt{fig/top.pdf}}}}
    \caption{Overview of the proposed multimodal fingertip and PROPRA.
 (a) Multimodal fingertip with tactile and proximity sensors.
 (b) Complementary sensing across manipulation phases: proximity before contact, tactile after contact, with proprioception available throughout and future action chunks providing additional sensorimotor context.
 (c) PROPRA independently aligns tactile and proximity histories. }
    \vspace{-2mm}
    \label{fig:top}
\end{figure}

To address this challenge, we investigate tactile and proximity sensing, which exhibit heterogeneous informative phases for imitation learning, and propose \underline{\textbf{PRO}}prioceptive-and-\underline{\textbf{PR}}oactive \underline{\textbf{A}}nchoring (PROPRA). This contrastive pretraining framework uses proprioception and action sequences, available throughout manipulation, as a common reference for learning fingertip sensor representations, where \emph{proactive} refers to action chunks that encode the robot's intended motion. By combining fingertip histories, current proprioception, and proactive actions, PROPRA temporally links fingertip perception with the robot's embodied state across past, present, and near-future interaction. Unlike image-anchored approaches, which rely on visual observations at a single time step, PROPRA uses proprioception and proactive actions that remain available throughout manipulation as anchors. This avoids forcing modalities that respond in different phases, while associating each modality with the states and motion during its informative phase. 

The main contributions of this work are threefold:
\begin{itemize}[leftmargin=*]
    \item We develop a fingertip sensing system that combines pre-contact proximity sensing with post-contact tactile sensing, and demonstrate its ability to capture complementary interaction states throughout the grasping process in real-robot experiments.
    \item We formulate the representation-learning challenge for complementary fingertip sensors that provide informative signals at different manipulation phases, and propose PROPRA, which uses proprioception and action chunks as a sensorimotor anchor. 
    \item We demonstrate higher average success rates than vision-only policies and image-anchored pretraining, while retaining pre-contact information in the learned representation, including on recordings from policy execution. 
\end{itemize}

% =============================================================================
\section{Related Work}
\label{sec:related_work}

Fingertip sensing provides direct information about physical interaction without relying solely on visual inference. Fundamental skills such as grasping and insertion particularly benefit from local perception, since vision-based position control can be unreliable when accurate interaction states must be estimated. We therefore review prior work on fingertip sensing and representation learning for multiple sensing modalities.

Tactile sensing has been widely used to improve robotic dexterity by directly measuring robot--object interaction~\cite{gelsight}. Vision-based tactile sensors observe contact-induced deformation of an elastic surface and are well suited to describing post-contact states such as object geometry, contact force, slip, and object pose during grasping~\cite{higuera2024sparsh, huang20243d, digit}. Alongside recent learning-based manipulation systems that generate actions from camera observations~\cite{pi05}, prior work has explored self-supervised visuotactile learning~\cite{sparshx}, shared representations across optical tactile sensors~\cite{anytouch}, and patch-level visual--tactile alignment~\cite{tacdino}.

However, tactile sensing alone cannot directly observe the interaction state before physical contact. Proximity sensing has therefore been used to estimate fingertip--object distance and relative configuration during the \emph{pre-grasp} or \emph{pre-shaping} phase~\cite{koyama2013preshape, proxtac,looktotouch}. Proximity sensing is informative before contact, whereas tactile sensing becomes informative after contact. Combining them thus provides complementary observations over a broader range of the grasping process.

This complementary sensing is also relevant to Learning from Demonstration (LfD), where fingertip information has been shown to improve manipulation learning~\cite{proxtac,sparshx,murooka2025TACT}. A key challenge is how to learn useful sensor representations from limited demonstrations. Contrastive pretraining provides one approach by exploiting relationships among observations to structure the latent space. Visual representation learning has used temporal relationships and goal states in demonstrations~\cite{r3m,vip}. For tactile sensing, exUMI~\cite{exumi} predicts future tactile observations conditioned on future actions, while $\tau$~\cite{tau} uses future visual observations as supervision. Visuotactile pretraining can also improve downstream policy performance~\cite{vital}. TactX~\cite{tactx} learns sensor-independent representations across optical, magnetic, and resistive tactile sensors, while HTT~\cite{htt} aligns optical and low-dimensional tactile observations through masked autoencoding and cross-modal alignment.

When multiple modalities must be represented jointly, a common strategy is to align them through a shared anchor. ImageBind~\cite{imagebind} uses images as an anchor and independently aligns other modalities to a shared latent space, while UniTouch~\cite{unitouch} applies a similar strategy to tactile representation learning. CentroBind~\cite{centrobind} further points out limitations of a fixed anchor modality, including limited anchor information, loss of modality-specific information, and insufficient modeling of correlations among non-anchor modalities.

Fingertip sensors are informative at different interaction phases, so direct same-time alignment may provide insufficient supervision. PROPRA instead uses proprioception and action sequences as a common sensorimotor anchor to learn complementary \emph{pre-grasp} and contact-related representations.

% =============================================================================
\section{Problem Formulation}
\label{sec:problem}

% まず，触覚と近接覚が操作の異なる局面で 有効となることを確認し， それが表現学習にもたらす課題を整理する．接触前に応答する近接覚と接触後に応答する触覚を同一の指先に組み合わせると，対象への接近から接触，保持に至る状態変化を途切れなく観測できる．接近中の距離と接触までの時間を近接覚が，接触の開始と把持状態を触覚が捉える．本研究では，触覚センサと近接センサを同一の指先に搭載した指先モジュールを構築した（図~\ref{fig:top}）．

% 触覚には感圧抵抗素子（FSR400S）を用いる．圧力に応じた抵抗変化を分圧回路で電圧に変換し，ボルテージフォロワを介して 3 チャネルの信号として読み出す．非接触時の出力は静止値に留まり，接触後に圧力に応じて変化する．近接覚には反射型の光学式距離検出回路を用いる．LED を数マイクロ秒間点灯させ，対象からの反射光をフォトトランジスタで受光し，その出力電圧をマイクロコントローラの AD 変換器で 6 チャネルで取得する．マイクロコントローラは 1~kHz 周期で各チャネルを順次AD変換し，生のAD値を 3~Mbps のシリアル通信で送信する．センサ信号は 100~Hz で取得し，軌道データと同期して保存する．素子はいずれも汎用の市販部品であり，既製の平行グリッパの指先に後付けできる．

We first examine how tactile and proximity sensing become informative at different stages of manipulation and clarify the resulting challenge for representation learning. By combining proximity sensing, which responds before contact, with tactile sensing, which becomes informative after contact, on the same fingertip, the interaction can be observed continuously from approach through contact to object holding. Proximity sensing captures the fingertip--object distance and the progression toward contact during approach, while tactile sensing captures contact onset and the subsequent grasp state. Based on this complementary sensing principle, we developed a fingertip module that integrates tactile and proximity sensors on the same fingertip (Fig.~\ref{fig:top}).

\subsection{Hardware design}
\label{subsec:hardware}

For tactile sensing, we use force-sensitive resistors (FSR400S). For proximity sensing, we use a reflective optical distance-sensing circuit. All sensor signals are acquired at 100~Hz and stored in synchronization with the robot trajectory. Both sensing elements are commercially available components and can be retrofitted to the fingertips of an existing parallel gripper.

\begin{table}[!t]
    \caption{Effect of naively adding fingertip sensing. Success rate [\%], mean $\pm$ SD over 3 seeds with 20 trials per seed. Sensor histories are concatenated to the observation without pretraining.}
    \label{tab:naive}
    \label{tab:policy}
    \centering
    \footnotesize
    \setlength{\tabcolsep}{4pt}
    \begin{tabular}{@{}lccr@{}}
    \toprule
    Task & Vision only & $+$ Tactile \& proximity & $\Delta$ \\
    \midrule
    PickCup    & 11.7\sd{7.6}  & 61.7\sd{5.8}  & $+$50.0 \\
    PickSponge & 35.0\sd{5.0}  & 51.7\sd{12.6} & $+$16.7 \\
    MovePen    & 31.7\sd{10.4} & 20.0\sd{8.7}  & $-$11.7 \\
    OpenLid    & 43.3\sd{7.6}  & 28.3\sd{15.3} & $-$15.0 \\
    \midrule
    Avg.       & 30.4 & 40.4 & $+$10.0 \\
    \bottomrule
    \end{tabular}
\end{table}

\subsection{Interval Structure of Sensor Responses}
\label{subsec:intervals}

Let $s^m_t$ denote the observation of modality $m\in{\mathrm{tac},\mathrm{prox}}$ at time $t$, with $q_t$ denoting the robot state and $a_t$ the action. We define the contact onset $t_c$ as the time at which the tactile signal rises from its stationary baseline; the detection criterion is described in Section~\ref{subsec:metrics}. The interval before $t_c$ is defined as \emph{pre-grasp}. Within this interval, the 60 control steps (approximately 2~s) immediately preceding $t_c$ are defined as \emph{approach}, and the earlier interval as \emph{pre}. The interval from $t_c$ onward is defined as \emph{contact}. These phase labels are used only for analysis and evaluation and are not provided during training.

The characteristics of the two sensor signals change across these phases. During \emph{pre}, both sensors remain near their stationary baselines. During \emph{approach}, the proximity signal begins to vary and becomes most discriminative immediately before contact, while the tactile signal remains near baseline. During \emph{contact}, the tactile signal rises in response to physical interaction. Thus, proximity sensing mainly covers the pre-contact interval, whereas tactile sensing becomes informative after contact. However, each sensor is discriminative only over a limited portion of the episode. In the demonstration data, at least one of the two sensors remains near its stationary value for more than half of the pretraining samples (Section~\ref{subsec:metrics}).

Having access to sensor information does not necessarily mean that a policy can use it effectively. When sensor histories are directly concatenated to a diffusion policy~\cite{dasari2024ditpi} that otherwise uses images and robot state, performance falls below the vision-only policy on some tasks (Table~\ref{tab:naive}). With limited demonstrations, the policy must learn both the meaning of each sensor signal and the interaction phase in which that signal should be used. This motivates learning sensor representations before policy training.

\subsection{Missing Supervision in Contrastive Pretraining}
\label{subsec:gap}
\label{subsec:anchor}

We next consider contrastive pretraining that aligns observations acquired at the same time as positive pairs. If one element of a pair remains constant over time, the loss does not require the encoder of the other element to preserve temporal distinctions. Because the informative intervals of tactile and proximity sensing only partially overlap, direct alignment yields supervision only in a limited portion of the trajectory. During \emph{approach}, the tactile signal remains near its stationary baseline even though the proximity signal becomes discriminative, so direct tactile--proximity alignment does not require the proximity encoder to distinguish states within this interval. Indeed, tactile observations cannot predict the corresponding proximity variation during \emph{approach} in our demonstration data (Section~\ref{subsec:metrics}). Removing degenerate samples does not resolve this issue: the remaining direct pairs are still concentrated near the contact transition, leaving most of the \emph{approach} phase without useful supervision.

From this observation, we require an anchor to satisfy the following three conditions:
\begin{enumerate}[label=(\roman*),leftmargin=*]
\item It is defined throughout manipulation, including intervals in which individual sensors do not respond.
\item It changes with the progress of manipulation and systematically covaries with the response intervals of the sensors.
\item It contains complementary sensorimotor context that is not a deterministic copy of any single sensor observation. 
\end{enumerate}
Condition (i) ensures that a common reference is available throughout the manipulation sequence, while useful supervision for each sensor remains concentrated in the phases where that sensor is informative; condition (ii) provides a meaningful relationship between the anchor and the interaction state, and condition (iii) prevents the anchor from reducing to an identity-like alignment with a single sensing modality. Tactile and proximity observations do not satisfy condition (i) when used as anchors for one another. Images satisfy condition (i), but their relationship to local physical quantities at the fingertips is indirect; whether condition (ii) holds depends on how clearly the fingertip--object interaction is observable in the image.

We therefore seek an anchor that is available throughout manipulation while remaining strongly coupled to the robot interaction state and motion. The next section introduces the sensorimotor anchor adopted in PROPRA to satisfy these requirements.

% ------------------------
\begin{figure*}[!tb]
    \centering
    \includegraphics[width=0.98\linewidth]{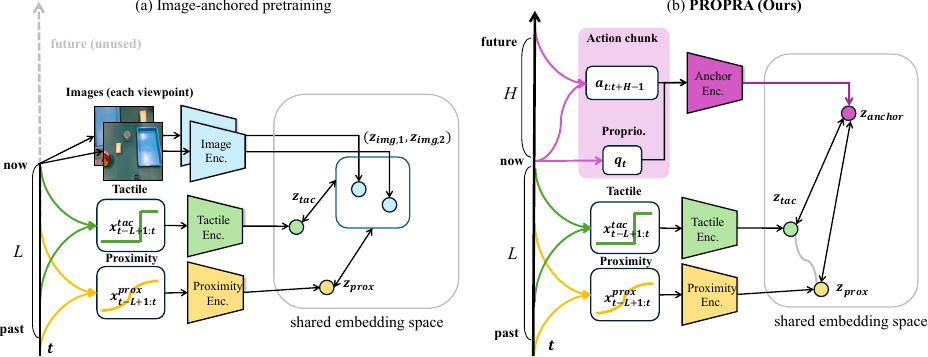}
    % {\fbox{\parbox[c][70mm][c]{0.98\columnwidth}{\centering\footnotesize
    % PLACEHOLDER: \texttt{fig/anchor.pdf}}}}
    \caption{\textbf{Pre-training for sparse-response sensors.} (a) Image-anchored alignment: sensor histories (past L steps) are aligned to image embeddings at time t; effective supervision concentrates in the brief contact window. (b) Our \textbf{PROPRA} aligns each sensor history to a temporally shared reference, the current state $q_t$, and the forthcoming action chunk $a(t:t+H-1)$, which is defined and varies at every $t$; no images are used. Dashed line in (b): no direct sensor-to-sensor objective — cross-sensor alignment emerges through the shared anchor.}
    
    %\vspace{-2mm}
    \label{fig:anchor}
    \end{figure*}
% -------------------------

% =============================================================================
\section{Method}
\label{sec:method}

% [v5] merged into Section IV-A to remove repetition

\subsection{Proprioceptive and Proactive Anchored Pretraining}
\label{subsec:overview}

PROPRA uses proprioception and action sequences as a common sensorimotor anchor for fingertip sensors whose informative signals arise at different stages of interaction, learns complementary representations of \emph{pre-grasp} and contact-related states, and transfers them to policy learning. It has two stages: pretraining of the sensor encoders (Fig.~\ref{fig:anchor}) and policy learning (Fig.~\ref{fig:policy}). In pretraining, a modality-specific encoder maps each sensor history to an embedding that is aligned independently with a common anchor embedding; sensor modalities are never paired directly, and their correspondence is formed only through the shared anchor. In policy learning, only the pretrained sensor encoders are transferred as observation encoders of the diffusion policy; the anchor and its encoder are used solely in pretraining.

For each sensor encoder, we use the most recent $L$-step history ending at time $t$. The anchor consists of the proprioceptive state at time $t$ and an $H$-step action chunk starting from $t$:
\begin{align}
x^m_t &= \big[\,s^m_{t-L+1},\,\dots,\,s^m_{t}\,\big]\in\mathbb{R}^{L\times d_m},
\label{eq:sensor-input}\\
\bar{a}_t &= \big[\,a_t^{\top},\,a_{t+1}^{\top},\,\dots,\,a_{t+H-1}^{\top}\,\big]^{\top}\in\mathbb{R}^{HA},
\label{eq:action-chunk}\\
h_t &= \big[\,q_t^{\top},\,\bar{a}_t^{\top}\,\big]^{\top}\in\mathbb{R}^{P+HA}.
\label{eq:anchor}
\end{align}
In our implementation, $q_t$ includes the end-effector pose and gripper opening ($P=8$), while $a_t$ includes seven commanded joint positions and one gripper command ($A=8$). With $H=10$, the anchor $h_t$ is therefore 88-dimensional.

The anchor $h_t$ satisfies the three requirements of Section~\ref{subsec:anchor}: it is defined over the entire trajectory because the robot always has a proprioceptive state and receives control commands (i); it covaries with the sensors' response phases, since the end effector moves toward the object before contact and the gripper command changes as the fingers close around contact (ii); and it is built entirely from robot-side variables, so it is not a deterministic copy of either tactile or proximity observations (iii).

The windows in Eqs.~(\ref{eq:sensor-input}) and~(\ref{eq:action-chunk}) are asymmetric around $t$: the sensor input covers the history up to $t$, whereas the anchor covers the current proprioceptive state and the actions from $t$ onward. PROPRA thus associates each sensor history with the current embodied state and subsequent motion rather than matching simultaneous observations, a predictive structure related to Contrastive Predictive Coding~\cite{cpc}. The positive pairs at each $t$ are $(h_t,x^m_t)$ for each modality $m$; no direct pair $(x^{\mathrm{tac}}_t,x^{\mathrm{prox}}_t)$ is used, and each additional modality adds a single anchor--modality term, so the framework scales linearly with the number of sensors.

\subsection{Encoders and Learning Objective}
\label{subsec:encoder}

Each modality with $d_m$ channels is normalized channel-wise by its mean and standard deviation; action chunks extending beyond an episode are padded by repeating the final action. Each sensor encoder $g_m$ is a four-layer pre-norm Transformer encoder: inputs are linearly projected to $d_{\mathrm{model}}=128$ with sinusoidal positional encoding, mean-pooled over time, and projected to a $512$-dimensional embedding. Encoder parameters are not shared, because the modalities differ in physical quantity and response characteristics. The anchor encoder $g_a$ is a multilayer perceptron. All embeddings are $L_2$-normalized,
$u^m_i=g_m(x^m_i)/\lVert g_m(x^m_i)\rVert$ and
$z_i=g_a(h_i)/\lVert g_a(h_i)\rVert$,
so that their inner product corresponds to cosine similarity.

Within a mini-batch of $B$ samples, $z_i$ and $u^m_i$ from the same time step form a positive pair, and the remaining $u^m_j$ ($j\neq i$) are negatives. For each modality, we use the symmetric InfoNCE objective~\cite{cpc, clip}, averaged over both directions:
\begin{align}
    \mathcal{L}
    &=\frac{1}{|\mathcal{M}|}
    \sum_{m\in\mathcal{M}}\mathcal{L}_m,
    \label{eq:loss}
    \\
    \mathcal{L}_m
    &=-\frac{1}{2B}\sum_{i=1}^{B}
    \Bigg[
    \log
    \frac{\exp(z_i^{\mathsf T}u^m_i/\tau)}
    {\sum_{j=1}^{B}\exp(z_i^{\mathsf T}u^m_j/\tau)}
    \nonumber\\
    &\qquad\qquad
    +\log
    \frac{\exp(z_i^{\mathsf T}u^m_i/\tau)}
    {\sum_{j=1}^{B}\exp(z_j^{\mathsf T}u^m_i/\tau)}
    \Bigg].
    \label{eq:loss-modality}
\end{align}
where $\tau$ is a learnable temperature. The tactile and proximity losses are computed independently against the same anchor and averaged; no term couples the two sensor modalities. We claim no novelty for the objective. Image--sensor and direct sensor--sensor alignment baselines are expressed in the same framework by weighting the corresponding alignment terms; $\mathrm{cw}$ denotes the weight of the direct tactile--proximity term ($\mathrm{cw}=0$: image anchor alone; $\mathrm{cw}=1$: with the direct pair).

\subsection{Integration with Policy Learning}
\label{subsec:policy}

We use a Transformer-based diffusion policy~\cite{diffusionpolicy,dasari2024ditpi} consisting of an observation tokenizer, an observation encoder, and a noise-prediction network that generates action chunks by reversing the diffusion process conditioned on the observations. Only the sensor encoders $g_m$ are transferred, with the sensor ordering and history length $L$ kept as in pretraining; they are fine-tuned end-to-end with the policy rather than frozen. Neither the anchor nor its encoder is transferred, so at inference the robot needs only images, proprioceptive state, and sensor histories.

\begin{figure}[!t]
    \centering
    \IfFileExists{fig/policy.pdf}{\includegraphics[width=0.99\columnwidth]{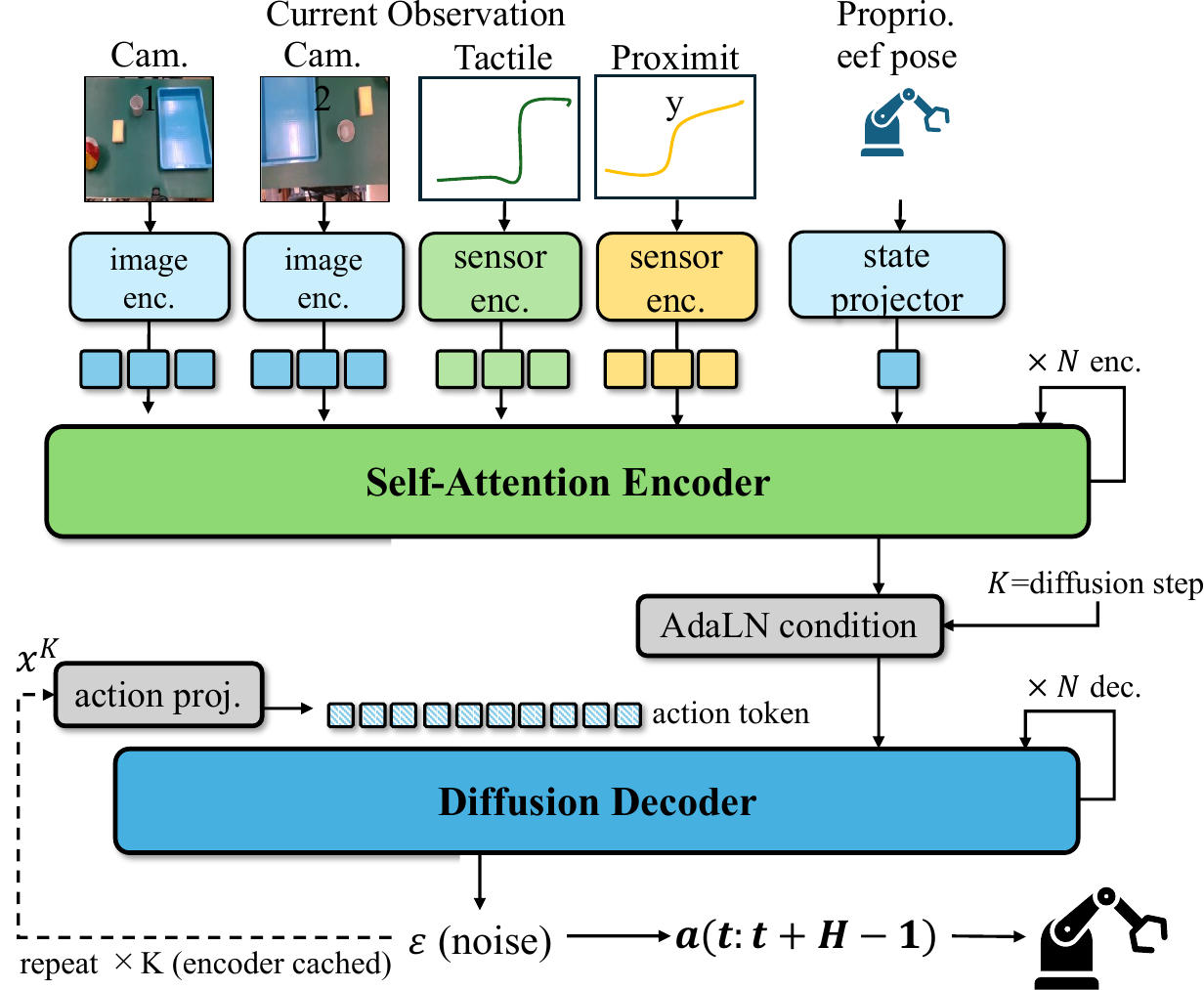}}
    {\fbox{\parbox[c][30mm][c]{0.9\columnwidth}{\centering\footnotesize
    PLACEHOLDER: \texttt{fig/policy.pdf}}}}
    \caption{Illustration of the model architecture for policy learning.}
    \vspace{-2mm}
    \label{fig:policy}
\end{figure}

% =============================================================================
\section{Experimental Setup}
\label{sec:setup}

\begin{figure}[!h]
    \centering
    \IfFileExists{fig/experiment.pdf}{\includegraphics[width=0.99\columnwidth]{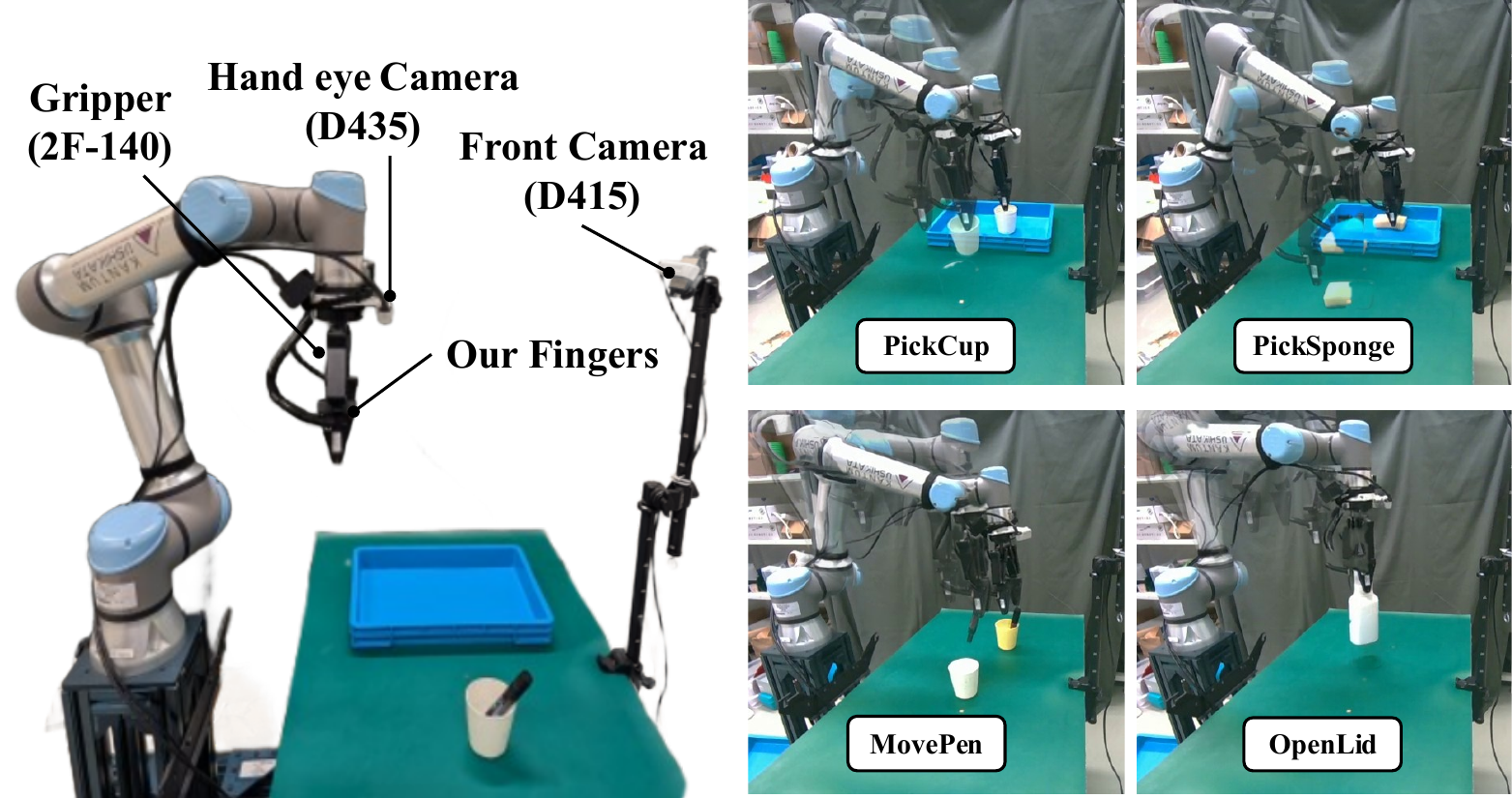}}
    {\fbox{\parbox[c][30mm][c]{0.92\columnwidth}{\centering\footnotesize
    PLACEHOLDER: \texttt{fig/experiment.pdf}}}}
    \caption {Experimental Setup. Our fingertip can observe tactile and proximity signals, while 2 RealSense cameras observe the scene (one camera is visible above; another is mounted on the end-effector).}
    \vspace{-2mm}
    \label{fig:robot}
\end{figure}

% 実機は UR5e に第\ref{subsec:module}節の指先モジュールを備えた平行グリッパを装着した構成である．指先センサは ROS~2 のトピックとして力 3 チャネルと距離 6 チャネルを配信し，計測は 100\,Hz，保存は 30\,Hz 周期である．方策の制御周期は 30\,Hz である．画像は手先（hand）と正面（front）の RGB-D カメラ 2 台から取得し，$224\times224$ にリサイズして用いる．エンドエフェクタ姿勢 6 次元とグリッパ開度 2 次元を身体状態とし，指令関節位置 7 次元と指令グリッパ開度 1 次元を行動とする．
The robot is a UR5e with a Robotiq 2F-140 gripper carrying the fingertip module of Section~\ref{subsec:hardware}, which publishes three force and six proximity channels as ROS~2 topics at 100~Hz, independently of the 30~Hz policy control loop; the most recent samples are buffered and stored with the robot state and action at every control step.
Two RealSense cameras (a wrist-mounted D435 and a workspace-mounted D415) provide a scene overview at $224\times224$ pixels. The robot state is a 6-D end-effector pose and a 2-D gripper opening; the action is seven commanded joint positions and one commanded gripper opening (Fig.~\ref{fig:robot}).

\subsection{Data Collection}
\label{subsec:tasks}

% PickCup，PickSponge，MovePen，OpenLid の 4 タスクを対象とする~\cite{tab:datasets}．PickCup と PickSponge は台上の対象を把持して持ち上げる．MovePen はペンを把持して容器へ移す．OpenLid は容器の蓋を把持して開ける．教示データは3D Mouseにより各タスク50エピソード収集し，物体の初期位置を毎回無作為化した．

We evaluate our method on four real-robot manipulation tasks: {PickCup}, {PickSponge}, {MovePen}, and {OpenLid}. In PickCup and PickSponge, shown in Fig.~\ref{fig:robot}, the robot grasps an object placed on a table and lifts it. In MovePen, the robot grasps a pen and transfers it into a container. In OpenLid, the robot grasps a container and opens its lid. For each task, we collect 50 demonstration episodes using a 3D mouse, randomizing the initial object position in every episode.

% \begin{table}[t]
%     \caption{Demonstration datasets used for evaluation ($T$: median episode length; $t_c$: median contact onset, both in control steps).}
%     \label{tab}
%     \centering
%     \small
%     \begin{tabular}{@{}lrrr@{}}
%     \toprule
%     Task & Episodes & $T$ & $t_c$ \\
%     \midrule
%     PickCup & 50 & 398 & 215 \\
%     PickSponge & 50 & 410 & 204 \\
%     MovePen & 50 & 381 & 196 \\
%     OpenLid & 50 & 432 & 228 \\
%     \bottomrule
%     \end{tabular}
% \end{table}

\subsection{Model Training}
\label{subsec:pretrain-setup}

Pretraining samples are obtained by subsampling the recorded sequence every five steps. Each sensor history $x^m_t$ contains the latest $L=50$ samples (0.5~s) of the 100~Hz sensor stream, read from the buffer at the corresponding control step so that it is aligned with the state and action of that step, while the action chunk $\bar{a}_t$ contains $H=10$ subsampled steps, covering approximately 1.67~s at the 30~Hz control rate. The anchor dimension is $P+HA=88$.

Mini-batches for the proposed condition are sampled non-uniformly: steps whose mean proximity value falls below the 20th percentile of its distribution, and steps at which the tactile amplitude crosses its 75th percentile in either direction, receive four times the sampling weight, and 25\% of each batch is drawn from the same interval of other episodes as hard negatives. The thresholds are computed from the evaluated sensors but are not used in the anchor. Image-anchored conditions use uniform sampling, so the difference from the proposed condition includes this sampling rule as part of the training recipe.

The image-anchored baseline uses an ImageNet-1K-pretrained ResNet-18 with batch normalization replaced by 16-group normalization. % [v5] encoder details already given in Section IV-B

All models are trained with AdamW (learning rate $10^{-4}$, weight decay $10^{-6}$, batch size 512) for 1000 epochs without data augmentation; 96\% of the episodes are used for training and the rest for validation, and the checkpoint with the lowest validation loss is selected. All conditions use the symmetric InfoNCE objective of Eq.~\eqref{eq:loss} with a learnable temperature $\tau$ initialized to $0.07$.
Each condition is trained with three seeds on one Intel Xeon Platinum 8558 CPU and one NVIDIA H200 GPU (141~GB VRAM). 
% data collection, training, and deployment use RoboManipBaselines~\cite{Murooka2026RMB}.

The policy (Fig.~\ref{fig:policy}) takes two camera images, the robot state, and sensor histories constructed as in pretraining, and is trained for 1000 epochs with batch size 64, 100 diffusion training steps, 8 inference denoising steps, and an action horizon of 16. Each of three seeds is evaluated over 20 trials; we report the mean success rate and standard deviation.

\subsection{Evaluation Metrics}
\label{subsec:metrics} 

\textbf{Contact onset and intervals.} The contact onset $t_c$ is defined as the first step at which any tactile channel deviates from its resting value by more than 150 AD units. The approach interval covers the 60 control steps (2~s at 30~Hz) preceding $t_c$. 

\textbf{Representation analysis.} Time-to-contact (TTC) is estimated from the frozen proximity embedding over the approach interval using ridge regression and reported as $R^2$. Cross-sensor retrieval evaluates whether a tactile embedding retrieves the proximity embedding from the same time step, reported as Recall@1 normalized by chance.

% =============================================================================

\section{Results and Discussion}
\label{sec:results}
 
% 我々は，提案する触覚・近接覚センサ構成と対照事前学習が，少数のデモンストレーションからの方策学習に与える効果を，実機ロボットの 4 タスクで評価する．特に，現在の身体状態とこれから実行する行動，すなわち過去・現在・未来をつなぐ身体側の信号に整合させるアンカー設計の有用性を示す。
 
% 比較対象として，以下の 2 種類の条件を用いた．事前学習の損失関数はすべて InfoNCE とした．
% \begin{enumerate}
%     \item \textbf{視覚のみ／事前学習なし：} 指先センサを用いない方策と，事前学習を行わずにセンサ履歴を観測へ連結した方策．前者はセンサ追加の効果を，後者は事前学習の必要性を確かめる基準である．
%     \item \textbf{画像アンカー：} ImageBind~\cite{imagebind} と同様に，各センサの表現を画像の表現へ整合させる事前学習．さらに，同時刻の触覚と近接覚を直接整合させる項を加えた条件（画像アンカー＋直接整合）も設けた．
% \end{enumerate}
% センサエンコーダの構造と学習設定は全条件で共通であり，ミニバッチの抽出規則のみ異なる（提案手法は接触付近のサンプルを優先して抽出する）．

% 評価は，方策の操作成功率と，事前学習したセンサ表現の解析の 2 段階で行う．表現の解析では，触覚がまだ反応しない接触前の区間に着目し，近接覚の表現から接触までの残り時間（time-to-contact; TTC）を線形回帰で推定できるかを調べる．推定の精度は決定係数 $R^2$（1 で完全に一致，0 で平均値による推定と同等）で表し，センサの生の値と，事前学習をしていないエンコーダを基準とする（手順は第\ref{subsec:metrics}節）．計算時間は事前学習で約 3--4 時間，方策学習で約 8 時間である．
% \begin{itemize}[leftmargin=*]
%   \item \textbf{RQ1:} 身体側の信号をアンカーとする事前学習は，画像アンカーより有効か（表~\ref{tab:anchor}）．
%   \item \textbf{RQ2:} 触覚と近接覚を直接対応付けて学習しなくても，2 つのセンサの表現は対応するか（図~\ref{fig:rq3-retrieval}）．
%   \item \textbf{RQ3:} 学習された表現は接触前の情報を保持し，方策の実行時に収録したデータでも保たれるか（表~\ref{tab:exp-ttc}，図~\ref{fig:rq4-readout}，\ref{fig:rq4-holdout}）．
% \end{itemize}

We evaluate the proposed tactile--proximity sensing and contrastive pretraining for policy learning from limited demonstrations on four real-robot tasks, focusing on whether aligning sensor histories with robot-side signals that connect the current embodied state to future actions is effective. We consider the following two types of comparison conditions. All conditions use the same InfoNCE objective (Section~\ref{subsec:pretrain-setup}).
\begin{enumerate}
\item \textbf{Vision only / no pretraining:} A vision-only policy without fingertip sensing, and a policy with sensor histories concatenated to the observations without pretraining; the former isolates the effect of adding fingertip sensing, the latter the need for pretraining.
\item \textbf{Image anchor:} Following VITaL~\cite{vital}, each sensor representation is aligned with an image representation, optionally with an additional direct alignment term between same-time tactile and proximity representations (\textbf{image anchor $+$ direct pair}, $\mathrm{cw}=1$).
\end{enumerate}
Encoder architecture and training settings are shared across conditions; only the mini-batch sampling differs, as the proposed method uses the non-uniform sampling of Section~\ref{subsec:pretrain-setup}.

We evaluate manipulation success rate and the pretrained representations. For the latter, we ask whether the remaining time to contact (TTC) can be linearly decoded from the proximity representation in the pre-contact interval, where the tactile sensor has not yet responded, and report the coefficient of determination $R^2$ ($1$: perfect prediction; $0$: predicting the mean), using raw sensor values and an untrained encoder as references (Section~\ref{subsec:metrics}). Pretraining takes about 3--4 hours and policy training about 8 hours.

We address the following research questions:
\begin{itemize}[leftmargin=*]
\item \textbf{RQ1:} Is pretraining with robot-side signals as an anchor more effective than image-anchored pretraining (Table~\ref{tab:anchor})?
\item \textbf{RQ2:} Can corresponding representations emerge between tactile and proximity sensing without directly aligning the two modalities during training (Fig.~\ref{fig:rq3-retrieval})?
\item \textbf{RQ3:} Do the learned representations preserve pre-contact information, and does this information remain available for data collected during policy execution (Table~\ref{tab:exp-ttc}, Figs.~\ref{fig:rq4-readout} and~\ref{fig:rq4-holdout})?
\end{itemize}

% -----------------------------------------------------------------------------
% -----------------------------------------------------------------------------
\begin{table*}[!t]
    \caption{Effect of pretraining and anchor choice with both fingertip sensors.
    Success rate [\%], mean $\pm$ SD over 3 seeds with 20 trials per seed.
    $+$ direct pair adds direct tactile--proximity alignment.
    Bold and underlined values indicate the best and second-best results in each
    column, respectively.}
    \label{tab:anchor}
    \centering
    \small
    \begin{tabular}{@{}lccccc@{}}
    \toprule
    Pretraining & PickCup & PickSponge & MovePen & OpenLid & Avg. \\
    \midrule
    None
    & 61.7\sd{5.8}
    & 51.7\sd{12.6}
    & 20.0\sd{8.7}
    & 28.3\sd{15.3}
    & 40.4 \\
    \midrule
    Image anchor
    & 75.0\sd{13.2}
    & \underline{75.0}\sd{13.2}
    & \underline{60.0}\sd{14.1}
    & \textbf{76.7}\sd{12.6}
    & 71.7 \\

    Image anchor $+$ pair 
    & \underline{76.7}\sd{7.6}
    & 71.7\sd{2.9}
    & \textbf{63.3}\sd{2.9}
    & \textbf{76.7}\sd{7.6}
    & \underline{72.1} \\
    \midrule
    \textbf{PROPRA (ours)}
    & \textbf{81.7}\sd{5.8}
    & \textbf{81.7}\sd{7.6}
    & \textbf{63.3}\sd{5.8}
    & \underline{73.3}\sd{5.8}
    & \textbf{75.0} \\
    \bottomrule
    \end{tabular}
\end{table*}

\subsection{RQ1: Does anchoring to past, present, and future body signals help, compared with image anchors and direct pairs?}
\label{subsec:rq1}

% 事前学習なしでセンサ履歴を連結した方策は，平均成功率を 30.4\,\% から 40.4\,\% に上げる一方，MovePen と OpenLid では視覚のみの方策を下回った（表~\ref{tab:naive}）．事前学習により，平均成功率はすべての条件で 70\,\% を超え，この 2 タスクも視覚のみの方策を上回る（表~\ref{tab:anchor}）．提案手法の平均は 75.0\,\% であり，画像アンカー（71.7\,\%）と画像アンカー＋直接整合（72.1\,\%）を上回った．タスク別では PickCup で 5.0 ポイント，PickSponge で 6.7 ポイント上回るが，これらの差は seed 間のばらつきより小さく，MovePen では同率，OpenLid では画像アンカーが 3.4 ポイント上回る．直接整合の有無による差は 0.4 ポイントであり，センサ同士を直接結び付けても成功率は上がらなかった．身体側の信号をアンカーとすることで，事前学習にカメラ画像を用いずに，平均で画像アンカーと同等以上の成功率が得られた．
Without pretraining, directly concatenating sensor histories increases the average success rate from 30.4\% to 40.4\%, but performance on MovePen and OpenLid remains below that of the vision-only policy (Table~\ref{tab:naive}). With pretraining, the average success rate exceeds 70\% for all conditions, and performance on these two tasks also surpasses the vision-only baseline (Table~\ref{tab:anchor}).

PROPRA achieves the highest average success rate (75.0\%, vs.\ 71.7\% for the image anchor and 72.1\% with the direct pair). It improves over the image anchor by 6.7 points on both PickCup and PickSponge (5.0 and 10.0 points over the direct-pair variant), although these differences are within the variation across seeds; MovePen is identical, and the image anchor is 3.4 points higher on OpenLid. Adding direct tactile--proximity alignment changes the average by only 0.4 points, so explicitly coupling the two representations brings no clear benefit.

Overall, robot-side anchoring matches or exceeds image-anchored pretraining on average without requiring camera images during pretraining.

% -----------------------------------------------------------------------------
\subsection{RQ2: Do tactile and proximity representations correspond without direct pairing?}
\label{subsec:rq2}

\begin{figure}[!t]
\centering
\IfFileExists{fig/fig_rq3_retrieval.pdf}{\includegraphics[width=\columnwidth]{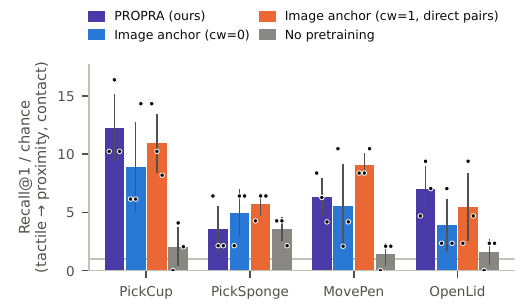}}
{\fbox{\parbox[c][30mm][c]{0.92\columnwidth}{\centering\footnotesize PLACEHOLDER: \texttt{fig/fig\_rq3\_retrieval.pdf}}}}
\caption{Finding the proximity embedding of the same time step from the tactile embedding during contact, on episodes not used for training. Accuracy is divided by chance (line at 1). Only Image anchor $+$ direct pair is trained on tactile--proximity pairs; No pretraining uses random weights. Bars: mean over three seeds; error bars: SD; dots: seeds. One correct retrieval changes a seed's value by 2.0--2.4.}
\vspace{-2mm}
\label{fig:rq3-retrieval}
\end{figure}

% 提案手法は触覚と近接覚を直接対応付けて学習しない．それでも 2 つのセンサの表現が対応するかを，学習に用いていないエピソードの接触中の各時刻について，触覚の表現から同じ時刻の近接覚の表現を探し当てる割合で確かめた（図~\ref{fig:rq3-retrieval}，偶然に当たる割合に対する倍率）．PickCup，MovePen，OpenLid では 6--12 倍となり，事前学習をしていないエンコーダ（1--2 倍）をすべての seed の組み合わせで上回った．2 つのセンサの対応は，共通のアンカーを介して間接的に形成された．一方，PickSponge では事前学習なしと同じ 3.6 倍にとどまり，画像アンカー＋直接整合は PickSponge と MovePen で提案手法を上回った．
Although PROPRA never aligns tactile and proximity representations directly, we test whether their correspondence emerges by retrieving, for each tactile embedding during contact in held-out episodes, the proximity embedding of the same time step, reported relative to chance (Fig.~\ref{fig:rq3-retrieval}).

On PickCup, MovePen, and OpenLid, PROPRA reaches 6--12$\times$ chance, above the untrained encoder (1--2$\times$) for every seed combination, so correspondence can emerge indirectly through the shared anchor. On PickSponge, however, PROPRA reaches only 3.6$\times$, comparable to the untrained encoder, and the image anchor with the direct pair exceeds PROPRA on PickSponge and MovePen.

% -----------------------------------------------------------------------------
\subsection{RQ3: Does the representation retain pre-contact information, also beyond the demonstration data?}
\label{subsec:rq3}

\begin{table}[!t]
\caption{Time-to-contact estimation before contact: $R^2$ ($\uparrow$) from the frozen proximity embedding, mean $\pm$ SD over 3 seeds. No pretr.: random weights. Raw: sensor values. Bold and underlined values indicate the best and second-best pretrained condition.}
\label{tab:exp-ttc}
\centering
\footnotesize
\setlength{\tabcolsep}{2pt}
\resizebox{\columnwidth}{!}{%
\begin{tabular}{@{}lccccc@{}}
\toprule
Task & PROPRA & Image & Image $+$ pair & No pretr. & Raw \\
\midrule
PickCup    & \textbf{0.694}\sd{0.002} & \underline{0.675}\sd{0.011} & 0.664\sd{0.005} & 0.535\sd{0.010} & 0.511 \\
PickSponge & \textbf{0.440}\sd{0.015} & \underline{0.377}\sd{0.034} & 0.372\sd{0.035} & 0.145\sd{0.016} & 0.327 \\
MovePen    & 0.544\sd{0.008} & \underline{0.545}\sd{0.010} & \textbf{0.554}\sd{0.007} & 0.500\sd{0.015} & 0.347 \\
OpenLid    & \textbf{0.680}\sd{0.026} & \underline{0.652}\sd{0.006} & 0.643\sd{0.005} & 0.622\sd{0.009} & 0.400 \\
\bottomrule
\end{tabular}}
\end{table}

\begin{figure}[!t]
\centering
\IfFileExists{fig/fig_rq4_ttc_readout.pdf}{\includegraphics[width=\columnwidth]{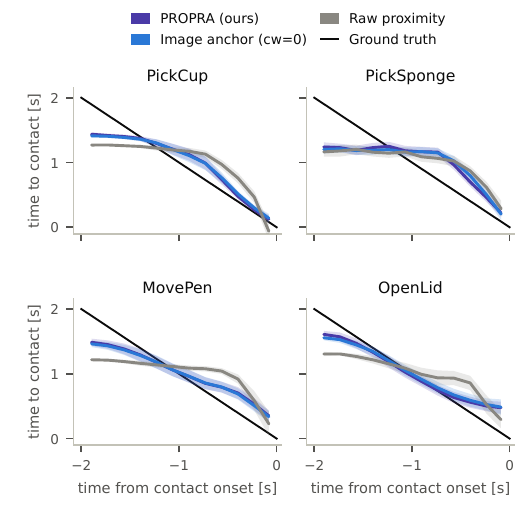}}
{\fbox{\parbox[c][30mm][c]{0.92\columnwidth}{\centering\footnotesize PLACEHOLDER: \texttt{fig/fig\_rq4\_ttc\_readout.pdf}}}}
\caption{Time to contact estimated from the frozen proximity embedding during the 2\,s before contact, averaged over the demonstration episodes (and over three seeds for the embeddings). Bands: 95\% CI across episodes. The diagonal is the true value.}
\vspace{-2mm}
\label{fig:rq4-readout}
\end{figure}

\begin{figure}[!t]
\centering
\IfFileExists{fig/fig_rq4_ttc_holdout.pdf}{\includegraphics[width=\columnwidth]{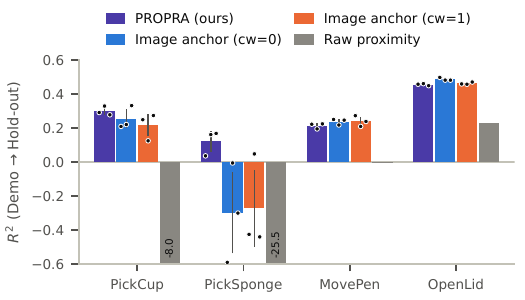}}
{\fbox{\parbox[c][30mm][c]{0.92\columnwidth}{\centering\footnotesize PLACEHOLDER: \texttt{fig/fig\_rq4\_ttc\_holdout.pdf}}}}
\caption{$R^2$ of the time-to-contact estimate on new recordings from policy execution, using the regression fitted on the demonstrations without refitting. Bars: mean over three seeds; error bars: SD; dots: seeds. Values below the axis are printed inside the bars.}
\vspace{-2mm}
\label{fig:rq4-holdout}
\end{figure}

% 接触前の 2 秒間について，近接覚の表現から接触までの残り時間を推定した（表~\ref{tab:exp-ttc}，図~\ref{fig:rq4-readout}）．事前学習した表現はいずれもセンサの生の値より正確に推定でき，提案手法は 4 タスク中 3 タスクで最も高い $R^2$ を示した．画像アンカーとの差が最大となったのは，生の値による推定が最も難しい PickSponge である（0.440 対 0.377）．MovePen では画像アンカー＋直接整合が最も高かった．また MovePen と OpenLid では，生の値に対する差（0.20，0.28）のうち 0.15 と 0.22 が事前学習をしていないエンコーダでも得られ，事前学習の効果が大きいのは PickCup と PickSponge である．触覚の表現からは，接触前の残り時間を推定できなかった（$R^2 \le 0.02$）．

% 次に，方策の実行中に新たに収録したデータ（各タスク 100 エピソード，事前学習にも方策学習にも用いていない）で同じ推定を行った．近接センサのゼロ点のずれは，各エピソード冒頭の静止値を教示データに揃えて補正した．このデータで推定器を学習し直すと，提案手法の $R^2$ は教示データ上の値の 65--90\,\%（95\,\% 信頼区間 $\pm 4$ 以内）を保ち，seed ごとの比較では画像アンカーの 2 条件を 24 回中 20 回上回った．教示データで学習した推定器をそのまま用いると（図~\ref{fig:rq4-holdout}），生の値では PickCup と PickSponge で $R^2$ が大きく負になった（$-8.0$，$-25.5$）のに対し，提案手法は全タスクの全 seed で正の値を保った．PickSponge では画像アンカーの 2 条件の平均が負となり，正を保ったのは提案手法のみである．一方，MovePen と OpenLid では画像アンカーの方が高かった．

Over the 2\,s before contact (Table~\ref{tab:exp-ttc}, Fig.~\ref{fig:rq4-readout}), all pretrained representations predict TTC more accurately than the raw signal, and PROPRA gives the highest $R^2$ on three of the four tasks; the largest gain over the image anchor is on PickSponge, where prediction from the raw signal is hardest (0.440 vs.\ 0.377), while the image anchor with the direct pair is highest on MovePen. On MovePen and OpenLid, most of the gain over the raw signal is already obtained without pretraining (0.15 of 0.20 and 0.22 of 0.28), so the effect of pretraining is concentrated on PickCup and PickSponge. The tactile representation does not predict TTC ($R^2 \le 0.02$).

On 100 additional episodes per task recorded during policy execution and used for neither pretraining nor policy learning, after aligning each episode's initial stationary proximity value with the demonstrations to compensate for zero-point drift, a linear predictor refit on these data retains 65--90\% of PROPRA's demonstration $R^2$ (95\% CI within $\pm4\%$), and PROPRA outperforms both image-anchor conditions in 20 of 24 seed-wise comparisons.

Applying the demonstration-fitted predictor without refitting (Fig.~\ref{fig:rq4-holdout}), the raw signal yields strongly negative $R^2$ on PickCup and PickSponge ($-8.0$ and $-25.5$), and both image-anchor conditions turn negative on PickSponge, whereas PROPRA stays positive on every task and seed; the image-anchor conditions are higher on MovePen and OpenLid.
 
\subsection{Limitations and future perspective}
\label{subsec:limits}

% 今回のような方策学習のケースにおいては、ロボット自身の姿勢と行動は追加のデータやラベルなしに実現できることがメリットであり、把持シーンにおける位相の変化に着目したセンサの構成はその利点が強調されたが、提案法におけるアンカー設計に関しては、従来型の画像ベースのアンカー設計に絶対的に有利でない場合も確認されている。例えば、OpenLid では画像アンカー（76.7\,\%）が提案手法（73.3\,\%）を上回る事例がある。このタスクは蓋を回す間に接触が繰り返される動作の特徴を有しており，指先の感覚の変化も大きいため，本研究が前提とする疎な応答の混在が成立しないため，教師データにおける画像とセンサの応答の変化の関係性が十分に計測されたことにあると考えられる．一方、この結果は，利得が疎な応答という前提の下で生じることを裏づける事例であるが，センサ情報が全区間において豊富である場合においても必ずしも性能を強化するわけでないことを示しており，今後の方針への課題である．

% 今回のFinger tipの設計と提案法の目的は，把持のシーンにおける特徴を仮定し，既製のグリッパに後付けした2種類のセンサと少数のデモンストレーションの模倣学習に向けられている。特定のデバイスを想定した大規模基盤モデルとの比較は，センサの構成の拘束条件と事前学習方針に着目して検討するため、今回の比較の対象としていない。一方で，今後の検討材料は、Sparsh-X~\cite{sparshx}のような多モーダル基盤モデルを、独自性のあるセンサの利用と結びつけることができるかという視点である．今回の手法のようにあるロボット操作を仮定し、対応するセンサ構成が検討された場合に，それに必要な追加学習として本手法が適用できると期待する．そうしたスケーラビリティとの関連性その検証は今後の課題である．

Robot proprioception and actions require no additional data collection or annotation in this setting, a practical advantage of sensorimotor anchors, especially for sensors designed to capture phase changes during grasping. The proposed anchor is not uniformly superior, however: on OpenLid the image anchor reaches 76.7\% versus 73.3\% for PROPRA. Repeated contact while rotating the lid produces substantial fingertip variation over a long portion of the trajectory, so the sparse-response assumption is less pronounced and the demonstrations may already contain enough correspondence between visual observations and sensor changes for image-based alignment. This is consistent with the advantage of PROPRA being most apparent when informative responses are temporally sparse, and it also shows that the anchor does not necessarily help when sensor information is rich over much of the trajectory, which remains an important limitation.

Our fingertip design and pretraining target a specific setting: grasp-oriented manipulation with two sensors retrofitted to an existing gripper and imitation learning from limited demonstrations. We therefore do not compare with large-scale foundation models built for particular tactile devices, since our focus is the sensing configuration and the pretraining objective. Combining PROPRA with multimodal foundation models such as Sparsh-X~\cite{sparshx}, for example as an adaptation stage when task-specific or non-standard sensors are introduced, and evaluating its scalability to broader sensing configurations, are left for future work.

% =============================================================================
\section{Conclusion}
\label{sec:conclusion}

% 本論文は，把持において方策が必要とする接触状態が外部カメラからも単一の指先センサからも直接には得られないという制約から出発した．応答区間が相補的な触覚と近接覚を指先に搭載したグリッパを構築し，これを方策の観測に連結するだけでは操作成功率が改善しないこと，およびその原因が各センサの応答区間の偏りにあることを実機 4タスクで示した．観測同士を整合させる対照事前学習が接触前の区間で教師を持たない状況に対して、3条件を満たすアンカーとして proprioception と行動チャンクを選び，事前学習手法 PROPRA を構成した．平均成功率は視覚のみの 30.4\,\%，指先センサの素朴な追加の 40.4\,\% に対し 75.0\,\% であり，画像をアンカーとする事前学習の 72.1\,\% を上回る．凍結表現の probe では，接触タイミングと接近区間の情報が画像をアンカーとする条件と同等以上に保持され，直接の対を用いずにセンサ間の対応が形成された．
This study addresses the difficulty of obtaining the contact states needed for grasping from either external cameras or a single fingertip sensor. We built a gripper with proximity and tactile sensors that are informative before and after contact, respectively, and proposed PROPRA for pretraining their representations. The results indicate that sensorimotor signals serve as useful anchors for sensors whose informative responses are confined to particular intervals, without direct alignment between observation modalities.
% 本研究では，事前学習において我々が提案したFingertipの前提として検証を行ったが，今後の課題として，汎化性をより拡張させるために，複数のツールや追加モダリティを用いた検討を行う予定です．特に，把持力やトルク，接触音の認識の導入を含むより多様なセンサ情報を通じて，提案された表現学習手法が堅牢性を持たせることを示すとともに，重要な検討事項の一つです．また，視覚情報を統合することで環境の全体的な状況を把握でき，局所的な近接センシングや触覚センシングを補完できる可能性があります．最後に，描画やバターを塗るといったより複雑なタスクに本手法を適用することは，その幅広い適応性を示す上で有望な方向性であると考えられます．
% 今後は，観測モダリティ同士ではなく身体側の信号をアンカーとする設計は，応答区間が限られる他のセンサ，たとえば接触音や振動への拡張が考えられる．
Future work will consider other tools and sensing modalities, including grasp force, torque, contact sound, and vibration, and test whether the learned representations remain effective across sensors with different physical characteristics and response patterns.

% =============================================================================
\ifblind
\else
\section*{Acknowledgment}
This work was supported in part by JST CREST, Japan (Grant Number JPMJCR2553) through the project ``MORAL: Morphoception-Oriented Reasoning and Action with Language,'' in part by the National Institute of Advanced Industrial Science and Technology (AIST) Policy Budget Project, ``Research and Development of Foundation Models for Generative AI in the Physical Domain,'' and in part by the ``Development Acceleration Use Program'' of ABCI 3.0 provided by AIST and AIST Solutions.
\fi

\bibliographystyle{IEEEtran}
\bibliography{reference}

% =============================================================================

\end{document}